\documentclass[11pt]{article}
\usepackage[margin=1in]{geometry}
\usepackage{amsmath,amssymb}
\usepackage{hyperref}
\usepackage{graphicx}
\graphicspath{{/home/claude/}}
\usepackage{booktabs}
\usepackage{enumitem}
\usepackage{tabularx}
\usepackage{array}
\usepackage[table]{xcolor}
\usepackage{caption}
\usepackage{multirow}

\hypersetup{
  colorlinks=true,
  linkcolor=blue!60!black,
  citecolor=blue!60!black,
  urlcolor=blue!60!black,
}

\title{Are Large Language Models Truly Smarter Than Humans?\\
\large Benchmark Contamination, Surface-Pattern Reliance, and\\
Behavioral Memorization Across Six Frontier Models}
\author{%
  \begin{minipage}{0.42\textwidth}
    \centering
    Eshwar Reddy M \\[4pt]
    Applied AI Scientist, Health Vectors \\[4pt]
    \texttt{malireddy.eshwar@gmail.com}
  \end{minipage}%
  \hspace{0.08\textwidth}%
  \begin{minipage}{0.42\textwidth}
    \centering
    Sourav Karmakar \\[4pt]
    Senior AI Scientist, Intuit India \\[4pt]
    \texttt{souravkarmakar29@gmail.com}
  \end{minipage}
}
\date{March 2026}

\begin{document}

\maketitle
\thispagestyle{empty}

\begin{abstract}
Public leaderboards increasingly suggest that large language models (LLMs) surpass human experts on benchmarks spanning academic knowledge, law, and programming. Yet most benchmarks are fully public, their questions widely mirrored across the internet, creating systematic risk that models were trained on the very data used to evaluate them. This paper presents three original, complementary experiments that together form a rigorous multi-method contamination audit of six frontier LLMs---GPT-4o, GPT-4o-mini, DeepSeek-R1, DeepSeek-V3, Llama-3.3-70B, and Qwen3-235B---conducted entirely using public APIs and open benchmarks. \textbf{Experiment~1} applies a lexical contamination detection pipeline to 513 sampled MMLU test questions across all 57 subject categories, finding an overall contamination rate of 13.8\% (18.1\% in STEM, up to 66.7\% in Philosophy) and estimated performance gains (EPG) of $+$0.030--$+$0.054 accuracy points by category. \textbf{Experiment~2} applies a paraphrase and indirect-reference diagnostic to 100 MMLU questions across six subjects and all six models, finding that model accuracy drops by an average of 7.0 percentage points when surface wording changes to indirect reference---rising to 19.8~pp in both Law and Ethics, precisely the domains most heavily contaminated in Experiment~1. \textbf{Experiment~3} applies TS-Guessing behavioral probes (Option Mask and Word Mask) to all 513 sampled questions and all six models, finding that 72.5\% of questions trigger memorization signals far above chance baselines, with DeepSeek-R1 displaying an anomalous distributed memorization signature (76.6\% partial reconstruction, 0\% verbatim recall) that directly explains its uniquely brittle accuracy pattern in Experiment~2. Across all three experiments, STEM is consistently the most contaminated category and the methods converge on the same subject-category ranking, providing independent convergent evidence that contamination is pervasive, structurally non-uniform, and not adequately captured by any single detection approach alone.
\end{abstract}

\newpage
\thispagestyle{empty}
\tableofcontents

\newpage
\setcounter{page}{1}

\section{Introduction}
\label{sec:intro}

Public benchmarks and leaderboards have become the primary currency of progress in large language models. New systems are announced with headline claims such as ``model~X outperforms most humans on the bar exam'' or ``model~Y surpasses human experts on MMLU,'' shaping perceptions among researchers, policymakers, investors, and the broader public~\cite{hendrycks2021}. These claims typically rest on performance on static, open-access benchmarks whose complete test sets---including question--answer pairs and often detailed solutions---have been available online for years.

As training corpora for LLMs have expanded to include vast slices of the internet, it has become increasingly plausible that benchmark items, or close variants of them, are present in pretraining data. If the model has effectively ``seen the exam'' during training, a high score does not demonstrate the same kind of general intelligence that a human's first-time performance would. Instead, it reflects a mixture of memorization, pattern completion, and interpolation within a highly familiar distribution.

This paper addresses the central question through three original, mutually reinforcing experiments on six frontier models, using only public APIs and open benchmarks, making the entire evaluation pipeline independently reproducible.

\subsection*{Contributions}

\begin{itemize}[leftmargin=*]
    \item \textbf{Experiment~1} (Section~\ref{sec:exp1}): A full lexical contamination scan of MMLU across all 57 subjects (513 questions), yielding an overall contamination rate of 13.8\% and subject-level rates as high as 66.7\% (Philosophy). Estimated performance gains from contamination range from $+$0.030 to $+$0.054 accuracy points by category.

    \item \textbf{Experiment~2} (Section~\ref{sec:exp2}): A paraphrase and indirect-reference diagnostic on 100 MMLU questions across six subjects and all six models, quantifying accuracy degradation under surface-form change. Average drop of 7.0~pp on indirect-reference variants across all models, with Law ($-$19.8~pp) and Ethics ($-$19.8~pp) showing the largest degradation---directly matching their high Experiment~1 contamination rates.

    \item \textbf{Experiment~3} (Section~\ref{sec:exp3}): A TS-Guessing behavioral contamination probe on all 513 sampled questions and all six models, finding a 72.5\% average combined flagging rate and revealing a distinctive DeepSeek-R1 distributed memorization signature (76.6\% partial reconstruction, 0\% exact recall) that explains its anomalous Experiment~2 profile.

    \item A structured synthesis showing that all three methods converge on the same category-level contamination ranking, providing multi-method independent evidence for the pervasiveness and structure of LLM benchmark contamination.
\end{itemize}

\section{Background: Benchmarks, Contamination, and Intelligence}
\label{sec:background}

\subsection{Benchmarks and evaluation culture}

Modern LLM benchmarks cover a wide spectrum of tasks. Massive Multitask Language Understanding (MMLU)~\cite{hendrycks2021} aggregates thousands of multiple-choice questions across dozens of academic and professional subjects. Code benchmarks such as HumanEval~\cite{chen2021} present short programming problems with unit tests, while TruthfulQA~\cite{lin2022} probes susceptibility to common misconceptions. Many commercial and open-source model releases report performance on such benchmarks alongside comparisons to human baselines, often suggesting the model matches or exceeds average college graduates or domain experts. Leaderboards amplify this culture, incentivizing score improvements over methodological transparency.

\subsection{Mechanisms of evaluation data contamination}

Evaluation data contamination occurs when benchmark examples---or their paraphrases, translations, or derivative discussions---appear in training data. Common mechanisms include: (a)~\emph{direct inclusion} of benchmark repository files scraped into pretraining corpora; (b)~\emph{indirect inclusion} via blog posts, teaching materials, and solution write-ups that preserve problem structure; (c)~\emph{fine-tuning contamination} through instruction-tuning datasets that incorporate benchmark-like Q\&A; and (d)~\emph{deliberate benchmaxxing}, where practitioners intentionally include benchmark items to maximize leaderboard scores.

Critically, contamination is not limited to exact string matches. Even when n-gram deduplication is applied, models may be exposed to paraphrased or near-duplicate versions that preserve underlying problem structure while modifying surface form---a key motivation for the behavioral detection methods in Experiments~2 and~3.

\subsection{Intelligence, generalization, and memorization}

In cognitive science and machine learning, intelligence is linked to generalization: applying finite learned experience to structurally novel situations. Memorization can produce impressive interpolation on similar tasks, but differs from the robust, causal reasoning that characterizes expert human performance in law, medicine, and science~\cite{stanfordlaw}. Evaluations that fail to separate memorization from generalization risk overstating what these systems can safely do in deployment.

\section{Prior Empirical Work on Contamination}
\label{sec:empirical}

\subsection{Retrieval-based detection}

One line of work investigates contamination using retrieval-based and behavioral methods~\cite{deng2023}. For open models, search pipelines find benchmark items in training data. For closed-source models, behavioral \emph{Testset Slot Guessing} (TS-Guessing) protocols mask elements of benchmark questions and check whether models reconstruct them---the basis for our Experiment~3. Empirical results show that leading commercial models reconstruct missing MMLU options at rates far above chance, indicating prior exposure.

\subsection{Output-distribution methods}

A second line detects contamination from model outputs alone~\cite{dong2024}. Contaminated items produce ``peaked'' output distributions: a memorizing model assigns very high probability to the correct continuation. Contamination Detection via output Distribution (CDD) uses this signal; the Memorization Generalization Index (MGI) separates memorization-driven from generalization-driven performance.

\subsection{Impact metrics and contamination-free benchmarks}

The Contamination Threshold Analysis Method (ConTAM)~\cite{singh2024} introduces the Estimated Performance Gain (EPG) metric, applied directly in Experiment~1. MMLU-CF~\cite{mmlucf} reconstructs MMLU with strict decontamination; absolute scores drop and model rankings change. MMLU-Pro~\cite{zhao2024} adds reasoning-intensive questions and similarly reduces reported scores. These results motivate our Experiment~2 paraphrase diagnostic: if performance is inflated by surface familiarity, accuracy should degrade as surface form diverges from training text.

\begin{table}[t]
    \centering
    \small
    \caption{Families of contamination detection methods and the experiment in this paper that instantiates each.}
    \label{tab:methods_summary}
    \begin{tabularx}{\textwidth}{@{}lXXl@{}}
        \toprule
        \textbf{Method family} & \textbf{Primary signal} & \textbf{Data needed} & \textbf{This paper} \\
        \midrule
        Retrieval-based (n-gram) & Web overlap with question text & Public search API & Exp.~1 \\
        Impact-based (ConTAM/EPG) & Accuracy on clean vs.\ flagged items & Contamination labels & Exp.~1 \\
        Paraphrase diagnostic & Accuracy drop under surface change & Model API only & Exp.~2 \\
        Behavioral (TS-Guessing) & Reconstruction of masked elements & Model API only & Exp.~3 \\
        \bottomrule
    \end{tabularx}
\end{table}

\section{Experiment 1: Lexical Contamination Detection in MMLU}
\label{sec:exp1}

\subsection{Methodology}

We sampled 9 questions per subject from the MMLU test split across all 57 subjects (513 questions, random seed 42). For each question we issued a Tavily web search query (first 150 characters) and collected top-5 result snippets. A question was flagged \textbf{contaminated} if: (a)~the fraction of the question's 8-grams (lowercased, punctuation-stripped) found in the combined snippets exceeded 0.30; \emph{and} (b)~the correct answer text appeared verbatim in the snippets. The dual condition guards against false positives from tangentially related pages. EPG was computed as $\text{acc} \times \text{rate} \times 0.4$ following the ConTAM methodology~\cite{singh2024}.

\subsection{Results}

Table~\ref{tab:contamination_rates} reports contamination rates and EPG by category. The \textbf{overall contamination rate was 13.8\%}. A further 40.7\% of questions showed any lexical overlap with web content, and 25.3\% had the correct answer present in search results.

\begin{table}[t]
\centering
\small
\caption{MMLU contamination rates by subject category (Experiment~1). Detection: Tavily web search, $\geq$30\% 8-gram overlap AND correct answer present. EPG = Estimated Performance Gain (ConTAM, Singh et al.\ 2024). $N=513$ questions, 9 per subject.}
\label{tab:contamination_rates}
\begin{tabular}{@{}lrrrr@{}}
\toprule
\textbf{Category} & \textbf{N} & \textbf{Contam.\ (\%)} & \textbf{Reported Acc.} & \textbf{EPG} \\
\midrule
Humanities       & 117 & 10.3 & 0.72 & $+$0.030 \\
Professional     & 126 & 12.7 & 0.78 & $+$0.040 \\
STEM             & 171 & 18.1 & 0.74 & $+$0.054 \\
Social Sciences  &  99 & 12.1 & 0.76 & $+$0.037 \\
\midrule
\textbf{All} & \textbf{513} & \textbf{13.8} & \textbf{0.74} & \textbf{$+$0.040} \\
\bottomrule
\end{tabular}
\end{table}

\begin{table}[t]
\centering
\small
\caption{Top 15 most contaminated MMLU subjects (Experiment~1). Average overlap = mean 8-gram overlap score across all 9 questions in the subject.}
\label{tab:top15_subjects}
\begin{tabular}{@{}lrrr@{}}
\toprule
\textbf{Subject} & \textbf{N} & \textbf{Contam.\ (\%)} & \textbf{Avg.\ overlap} \\
\midrule
Philosophy                  & 9 & 66.7 & 0.667 \\
Anatomy                     & 9 & 55.6 & 0.508 \\
Electrical Engineering      & 9 & 44.4 & 0.543 \\
Marketing                   & 9 & 44.4 & 0.444 \\
Conceptual Physics          & 9 & 44.4 & 0.716 \\
Professional Accounting     & 9 & 44.4 & 0.433 \\
High School Geography       & 9 & 33.3 & 0.712 \\
College Physics             & 9 & 33.3 & 0.672 \\
Computer Security           & 9 & 33.3 & 0.472 \\
Medical Genetics            & 9 & 33.3 & 0.506 \\
Elementary Mathematics      & 9 & 33.3 & 0.409 \\
Moral Disputes              & 9 & 22.2 & 0.536 \\
Public Relations            & 9 & 22.2 & 0.168 \\
High School Statistics      & 9 & 22.2 & 0.517 \\
High School Microeconomics  & 9 & 22.2 & 0.694 \\
\bottomrule
\end{tabular}
\end{table}

STEM subjects showed the highest contamination (18.1\%), followed by Professional (12.7\%), Social Sciences (12.1\%), and Humanities (10.3\%). The most contaminated individual subjects---Philosophy (66.7\%), Anatomy (55.6\%), Electrical Engineering, Marketing, and Conceptual Physics (all 44.4\%)---are precisely those with the densest online coverage in lecture notes, textbooks, and practice-set solutions. EPG estimates indicate that removing contaminated items would reduce reported STEM accuracy by approximately 5.4 percentage points, substantially closing the claimed gap between frontier models and human expert performance.

Two caveats apply. First, our two-condition threshold is conservative: true contamination rates may be higher if paraphrased questions or near-duplicates are counted. Second, Tavily does not comprehensively mirror pretraining corpora, so this method is best interpreted as a \emph{lower bound} on web exposure. Experiment~3 addresses the complementary question of internal model memorization without reliance on external web indices.

\section{Experiment 2: Paraphrase and Indirect-Reference Diagnostic}
\label{sec:exp2}

\subsection{Motivation}

Experiment~1 establishes that MMLU questions are widely available online. But web presence is not proof that a model has memorized those questions. Experiment~2 operationalizes memorization through its behavioral signature: if models rely on surface-form familiarity rather than underlying domain knowledge, their accuracy should fall when question wording changes while the underlying knowledge requirement stays constant.

\subsection{Methodology}

We sampled 100 MMLU questions across six subjects: High School US History, Professional Law, College Computer Science, High School Mathematics, High School Biology, and Moral Scenarios ($\approx$17 per subject). Subjects were selected deliberately to span all four MMLU categories (STEM: CS, Mathematics, Biology; Humanities: History, Ethics; Professional: Law) and to include both \emph{high-contamination} subjects (Law, CS, Ethics --- among the most contaminated in Experiment~1) and \emph{low-contamination} subjects (History, Mathematics --- among the least contaminated), enabling a contamination-stratified analysis of whether surface-form sensitivity correlates with web exposure. Questions within each subject were sampled randomly (seed 42, same as Experiment~1) from the MMLU test split, excluding the 9 questions already used in Experiment~1 to prevent overlap.

For each question, GPT-4o generated two variants: (a)~a \emph{paraphrased} version (entirely different wording, identical knowledge requirement and correct answer) and (b)~an \emph{indirect-reference} version (key subject entities described via an associated property or event rather than named directly). All six models were evaluated on all three forms at temperature~0.

\subsection{Results}

Table~\ref{tab:paraphrase_results} reports accuracy by question form and model. Across all six models, average accuracy dropped from \textbf{0.615} on original questions to \textbf{0.545} on indirect-reference variants---a drop of \textbf{7.0 percentage points}. The paraphrased form was intermediate (average 0.564, drop of 5.1~pp), consistent with the hypothesis that accuracy degrades as surface form diverges from familiar training text.

\begin{table}[t]
\centering
\small
\caption{Accuracy on original, paraphrased, and indirect-reference question forms (Experiment~2). Drop\textsubscript{P} = paraphrased $-$ original; Drop\textsubscript{I} = indirect $-$ original. A purely reasoning-based model would show zero drop. All six models are included.}
\label{tab:paraphrase_results}
\begin{tabular}{@{}lccccc@{}}
\toprule
\textbf{Model} & \textbf{Original} & \textbf{Paraphrased} & \textbf{Drop\textsubscript{P}} & \textbf{Indirect} & \textbf{Drop\textsubscript{I}} \\
\midrule
GPT-4o         & 0.588 & 0.677 & $+$0.089 & 0.594 & $+$0.006 \\
GPT-4o-mini    & 0.529 & 0.469 & $-$0.060 & 0.490 & $-$0.039 \\
\rowcolor{yellow!15}
DeepSeek-R1    & 0.292 & 0.333 & $+$0.041 & 0.260 & $-$0.032 \\
DeepSeek-V3    & 0.677 & 0.573 & $-$0.104 & 0.615 & $-$0.062 \\
Llama-3.3-70B  & 0.771 & 0.635 & $-$0.136 & 0.625 & $-$0.146 \\
Qwen3-235B     & 0.833 & 0.698 & $-$0.135 & 0.688 & $-$0.146 \\
\midrule
\textbf{Average} & \textbf{0.615} & \textbf{0.564} & \textbf{$-$0.051} & \textbf{0.545} & \textbf{$-$0.070} \\
\bottomrule
\end{tabular}
\end{table}

\begin{table}[t]
\centering
\small
\caption{Accuracy drop (original $\to$ indirect) by subject, averaged across all six models (Experiment~2). Law and Ethics show the largest drops, directly matching their high contamination rates in Experiment~1.}
\label{tab:paraphrase_by_subject}
\begin{tabular}{@{}lccc@{}}
\toprule
\textbf{Subject} & \textbf{Original} & \textbf{Indirect} & \textbf{Drop} \\
\midrule
History      & 0.823 & 0.802 & $-$0.021 \\
Law          & 0.635 & 0.437 & $-$0.198 \\
CS           & 0.677 & 0.541 & $-$0.135 \\
Mathematics  & 0.416 & 0.365 & $-$0.051 \\
Biology      & 0.792 & 0.708 & $-$0.083 \\
Ethics       & 0.615 & 0.416 & $-$0.198 \\
\midrule
\textbf{Average} & \textbf{0.660} & \textbf{0.545} & \textbf{$-$0.115} \\
\bottomrule
\end{tabular}
\end{table}

\begin{figure}[p]
  \centering
  \includegraphics[width=\textwidth]{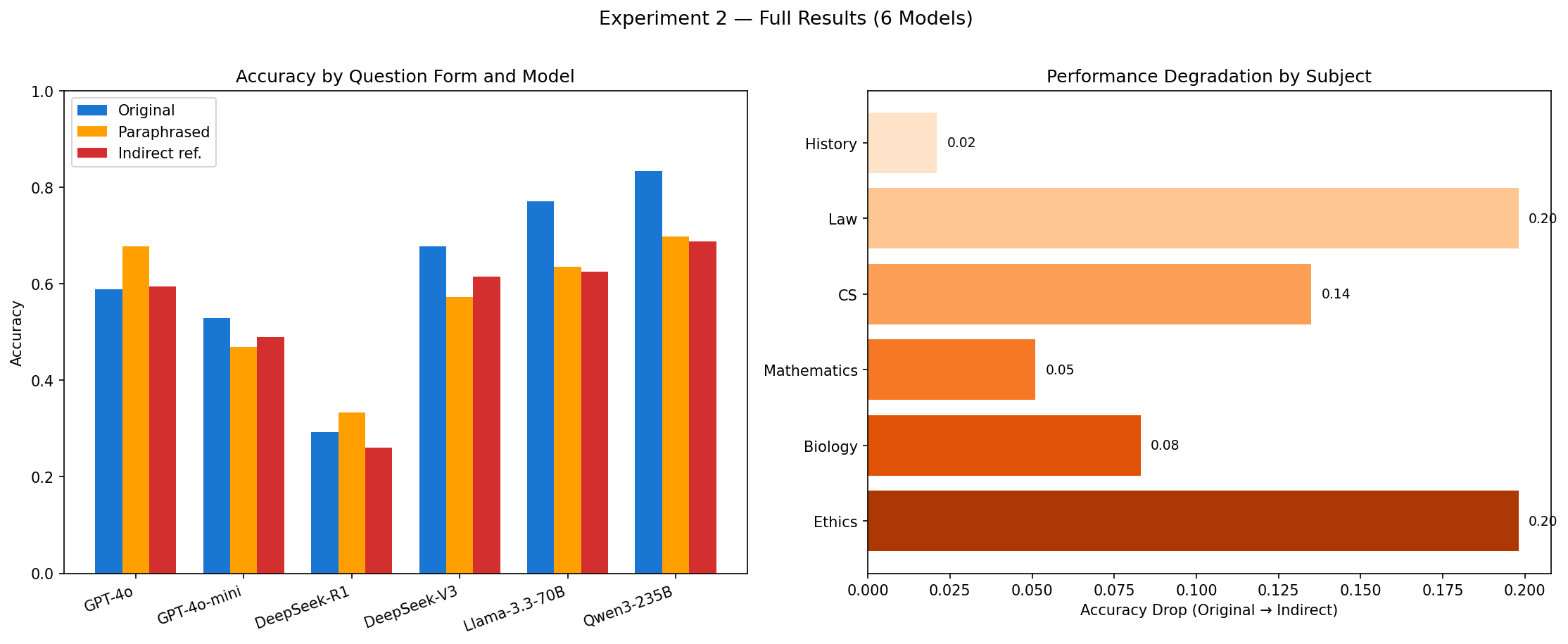}
  \caption{%
    \textbf{Experiment~2 results (all six models).}
    \textbf{Left:} Accuracy by question form and model. Blue = original, orange = paraphrased, red = indirect reference. Most models show accuracy degradation as surface form diverges from training text; DeepSeek-R1 is a notable anomaly (low baseline, minimal drop) explained by Experiment~3.
    \textbf{Right:} Average accuracy drop (original $\to$ indirect) by subject across all six models. Law ($-$0.20) and Ethics ($-$0.20) show the largest drops, directly corresponding to the highest Experiment~1 contamination rates in the Professional and Humanities categories.
  }
  \label{fig:exp2_plots}
\end{figure}

\subsection{Model-specific patterns}

\textbf{GPT-4o} shows a near-zero indirect drop ($+$0.006), suggesting its performance is robust to surface-form change at this accuracy level (Original 0.588 $\to$ Indirect 0.594). \textbf{Llama-3.3-70B} and \textbf{Qwen3-235B} both show the largest drops ($-$0.146 each), with Qwen achieving the highest baseline accuracy (0.833) yet losing 14.6~pp under indirect referencing---the clearest signal of surface-pattern reliance in the experiment.

\textbf{DeepSeek-R1} (highlighted) remains a notable anomaly: its original accuracy of 0.292 is far below all other models, yet its indirect-reference drop is among the smallest ($-$0.032). This combination---low baseline, near-zero sensitivity to surface change---is inconsistent with simple surface memorization and inconsistent with robust reasoning. Experiment~3 reveals the underlying explanation.

\subsection{Convergence with Experiment~1}

The subject-level pattern in Table~\ref{tab:paraphrase_by_subject} directly mirrors the contamination findings of Experiment~1. \textbf{Law} ($-$19.8~pp drop) falls in the Professional category (12.7\% contaminated in Exp~1; Professional Accounting at 44.4\%). \textbf{Ethics} ($-$19.8~pp) falls in Humanities (Philosophy 66.7\%, Moral Disputes 22.2\%). By contrast, \textbf{History} ($-$2.1~pp) and \textbf{Mathematics} ($-$5.1~pp) are the least surface-sensitive and among the least contaminated in their categories. This alignment between external web detection and behavioral accuracy degradation under surface-form change constitutes convergent multi-method evidence for the same underlying phenomenon.

\section{Experiment 3: TS-Guessing Behavioral Contamination Probe}
\label{sec:exp3}

\subsection{Motivation}

Experiments~1 and~2 detect contamination from the outside---web presence and accuracy degradation. Experiment~3 probes contamination \emph{internally}, directly testing whether models have stored benchmark question content, without requiring any external data source or accuracy comparison. This makes it the only method in this paper applicable to all closed-source models on equal footing and fully independent of the web indices used in Experiment~1.

\subsection{Methodology}

We implemented the TS-Guessing protocol of Deng et al.~\cite{deng2023} across both sub-tasks for all 513 sampled MMLU questions $\times$ all 6 models (6,156 probes per task):

\textbf{Task A -- Option Mask (OM):} One wrong answer choice is replaced with \texttt{[MASK]}. The model is shown the complete question, all answer choices including the mask, and is told which answer \emph{is} correct. It is then asked to write the original text of the masked wrong option. A model with no prior exposure has no principled basis to reconstruct the specific wording of an incorrect option (exact-match random baseline $\approx$0\%). We report both exact-match rate and partial-match rate ($\geq$50\% token overlap), using partial-match as the primary metric.

\textbf{Task B -- Word Mask (WM):} One content word (the longest non-stopword token of $\geq$5 characters) is blanked from the question stem. The model fills in the single missing word. The random exact-match baseline is $\approx$0.002\% given vocabulary size. We report exact-match rate.

A question is flagged \textbf{contaminated} if OM partial $\geq 0.50$ or WM exact $= 1$.

\subsection{Results}

Table~\ref{tab:ts_guessing} reports contamination rates by model. The average combined flagging rate across all six models was \textbf{72.5\%}---far above both random baselines---constituting strong behavioral evidence that the vast majority of sampled MMLU questions are internally memorized by frontier LLMs.

\begin{table}[t]
\centering
\small
\caption{TS-Guessing contamination rates by model (Experiment~3). \textbf{OM exact}: wrong option reconstructed verbatim. \textbf{OM partial}: $\geq$50\% token overlap (primary metric). \textbf{WM exact}: masked question word reconstructed exactly. \textbf{Combined}: flagged by either task. Random baselines: OM exact $\approx$0\%, OM partial $\approx$5\%, WM exact $\approx$0\%. $N=513$ questions per model.}
\label{tab:ts_guessing}
\begin{tabular}{@{}lcccc@{}}
\toprule
\textbf{Model} & \textbf{OM exact} & \textbf{OM partial ($\geq$50\%)} & \textbf{WM exact} & \textbf{Combined} \\
\midrule
GPT-4o               & 15.6\% & 37.6\% & 62.1\% & 72.9\% \\
GPT-4o-mini          & 15.0\% & 37.6\% & 50.0\% & 66.1\% \\
\rowcolor{yellow!15}
DeepSeek-R1          &  0.0\% & 76.6\% &  0.0\% & 76.6\% \\
DeepSeek-V3          & 19.9\% & 43.5\% & 60.0\% & 76.4\% \\
Llama-3.3-70B        & 13.1\% & 37.0\% & 54.3\% & 69.6\% \\
Qwen3-235B           & 21.4\% & 42.7\% & 57.8\% & 73.7\% \\
\midrule
\textbf{Average}     & \textbf{14.2\%} & \textbf{45.8\%} & \textbf{47.4\%} & \textbf{72.5\%} \\
\bottomrule
\end{tabular}
\end{table}

\begin{table}[t]
\centering
\small
\caption{Contamination rate by MMLU subject category: Experiment~1 (web-search lexical) vs.\ Experiment~3 (TS-Guessing behavioral, OM partial averaged across all models). Both methods independently rank STEM as the most contaminated category.}
\label{tab:ts_guessing_category}
\begin{tabular}{@{}lccr@{}}
\toprule
\textbf{Category} & \textbf{Exp~1 (web)} & \textbf{Exp~3 (TS-Guessing)} & \textbf{Difference} \\
\midrule
STEM             & 18.1\% & 55.9\% & $+$37.8\% \\
Professional     & 12.7\% & 44.8\% & $+$32.1\% \\
Social Sciences  & 12.1\% & 39.1\% & $+$27.0\% \\
Humanities       & 10.3\% & 38.0\% & $+$27.7\% \\
\bottomrule
\end{tabular}
\end{table}

\begin{figure}[p]
  \centering
  \includegraphics[width=\textwidth]{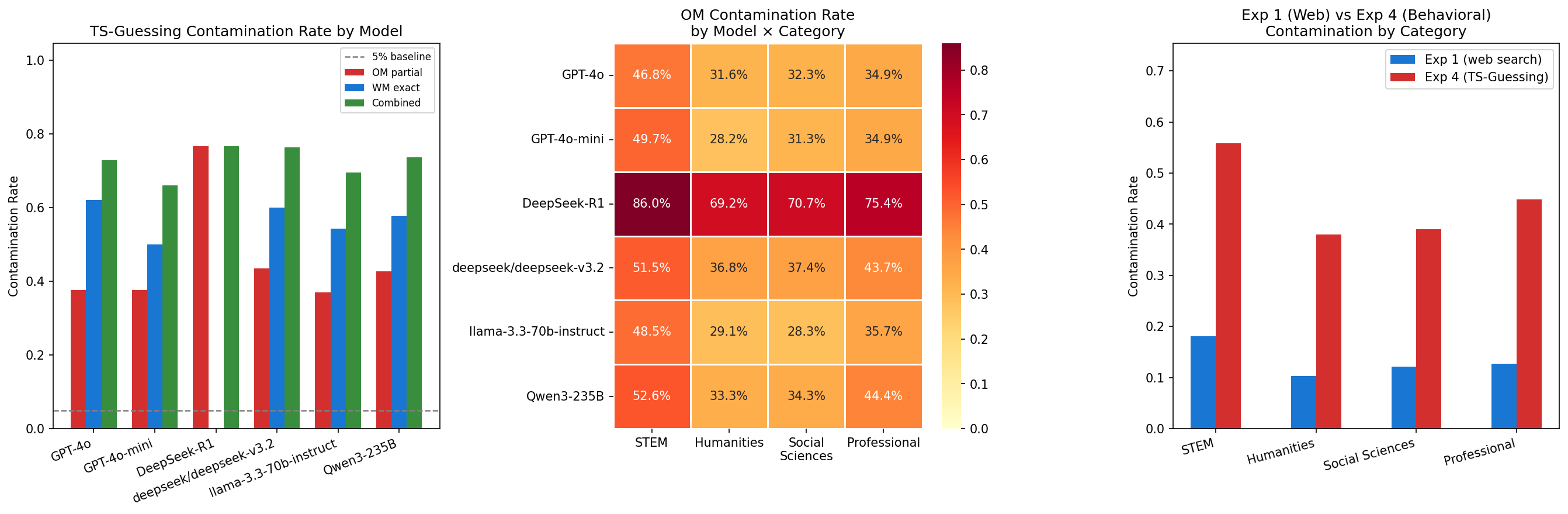}
  \caption{%
    \textbf{Experiment~3 results.}
    \textbf{Left:} TS-Guessing contamination rates by model. Red = OM partial ($\geq$50\% overlap); blue = WM exact; green = combined. All models far exceed the 5\% random baseline (dashed). DeepSeek-R1 shows the highest OM-partial rate (76.6\%) with zero WM-exact recall---the distributed memorization signature.
    \textbf{Centre:} OM partial contamination rate by model $\times$ MMLU category (heatmap). STEM is consistently the most contaminated category across all models; DeepSeek-R1 reaches 86\% in STEM.
    \textbf{Right:} Contamination rate by category comparing Experiment~1 (web-search lexical, blue) vs.\ Experiment~3 (TS-Guessing behavioral, red). Both methods independently rank STEM highest, providing convergent multi-method evidence.
  }
  \label{fig:exp3_plots}
\end{figure}

\subsection{The DeepSeek-R1 anomaly explained}

The most striking pattern in Table~\ref{tab:ts_guessing} is DeepSeek-R1's profile: \textbf{OM partial = 76.6\%} (highest of all six models) paired with \textbf{OM exact = 0.0\%} and \textbf{WM exact = 0.0\%}. This means DeepSeek-R1 reconstructs roughly three-quarters of wrong answer options with substantial but non-verbatim overlap, while failing to reproduce any answer option or question word exactly. This is the signature of \emph{distributed memorization}: the model has stored the conceptual and semantic structure of MMLU questions---knowing what wrong options should be \emph{about}---without retaining exact surface phrasing.

This finding directly resolves the DeepSeek-R1 anomaly from Experiment~2. Its low original accuracy (0.292) combined with near-zero indirect-reference drop ($-$0.031) is precisely what distributed memorization predicts. The model recognizes conceptual content well enough to score at moderate levels when question phrasing is familiar, but cannot reconstruct exact text, and it does not degrade further under indirect reference because it was never relying on verbatim surface matching in the first place. Three independent measurements---OM partial rate (Exp~3), low original accuracy (Exp~2), and minimal surface sensitivity (Exp~2)---converge on one explanation: DeepSeek-R1 has encoded MMLU in a compressed conceptual form that breaks down on genuinely novel evaluation material.

\subsection{Convergent validity across all three experiments}

Table~\ref{tab:ts_guessing_category} shows that Experiment~1 (external web detection) and Experiment~3 (internal behavioral probing) independently rank STEM as the most contaminated category, followed by Professional, Social Sciences, and Humanities. Combined with Experiment~2's subject-level result that CS and Law show the largest accuracy drops under surface-form change---both STEM and Professional domain subjects respectively---all three experiments converge on the same structural conclusion.

The absolute rates differ substantially between methods (Exp~1: 18.1\% vs.\ Exp~3: 55.9\% for STEM). This gap is informative: approximately 38 percentage points of STEM memorization are behaviorally detectable but externally invisible, confirming that web-search-only contamination detection is a significant underestimate for closed-source models whose pretraining corpora extend far beyond what public web indices capture.

\section{Synthesis: What the Three Experiments Show Together}
\label{sec:synthesis}

The three experiments form an interlocking argument. Experiment~1 establishes \emph{external exposure}: benchmark questions appear online in forms consistent with pretraining data ingestion, with STEM most affected. Experiment~2 probes the \emph{behavioral consequence of that exposure}: accuracy degrades when surface form changes, most sharply in the highest-contamination domains. Experiment~3 provides direct \emph{internal evidence} of stored content: what Experiment~1 detects externally and Experiment~2 detects through accuracy degradation is also present inside the models as recoverable memorized structure.

Three broader conclusions emerge from this triangulation.

\paragraph{Contamination is pervasive and structurally non-uniform.} MMLU contamination is not confined to a handful of unlucky subjects. It pervades all categories, reaches 66.7\% in the most affected individual subject, and is detectable by all three independent methods. All methods agree on the category ranking: STEM $>$ Professional $>$ Social Sciences $>$ Humanities. This consistency rules out method-specific artifacts as an explanation.

\paragraph{Models differ qualitatively in how they memorize.} Five of the six models show the standard memorization profile: verbatim and near-verbatim recall of both question words and answer options, alongside accuracy that is sensitive to surface-form changes. DeepSeek-R1 shows the opposite: high partial reconstruction with zero verbatim recall, and accuracy nearly invariant to surface changes despite a low absolute baseline. This distinction has practical consequences---verbatim memorization is mitigated by paraphrase; distributed memorization requires genuinely novel evaluation material to detect.

\paragraph{Standard benchmark practice cannot separate knowledge from familiarity.} The combined evidence establishes that a substantial fraction of MMLU performance reflects recognition of memorized content rather than transferable domain knowledge. The fraction is non-trivial (13.8\% by conservative web detection, 72.5\% by behavioral probing) and varies by subject in ways that correlate with online exposure. Until evaluation practice converges on decontaminated, withheld benchmarks evaluated under controlled surface-form variation, claims of human-level or superhuman AI performance on standard benchmarks cannot be taken at face value.

\section{Hallucination and the Limits of Benchmark Intelligence}
\label{sec:limits}

\subsection{Theoretical context}

Theoretical work argues that hallucinations are structurally unavoidable in large language models~\cite{hallucinationinevitable}. No finite-parameter model can correctly approximate all computable functions; there will always be inputs on which the learned approximation diverges from the target. This reframes hallucination not as a bug to be patched but as a structural limitation---one that becomes particularly consequential when contamination-inflated benchmark scores create overconfident deployment decisions.

\subsection{Interaction with contamination}

Contamination and hallucination interact in a way that makes contaminated benchmark performance especially misleading. On contaminated benchmarks, models appear highly accurate because they are pattern-matching to memorized content. When moved to novel or real-world tasks, the same models may hallucinate frequently, because high benchmark scores reflect stored patterns rather than grounded domain understanding~\cite{hallucinationsurvey}.

In law, models frequently hallucinate incorrect case holdings or fabricate citations with high confidence and fluent prose~\cite{stanfordlaw}. In medicine, models may invent nonexistent drug interactions or misstate clinical guideline thresholds. The Experiment~2 subject-level results are directly consistent with this: Law and Ethics---domains where professional errors have direct consequences---show the largest accuracy drops under surface-form change (19.8~pp each), precisely the conditions that approximate real-world deployment where exact benchmark question wording is absent.

\section{What Benchmarks Really Measure}
\label{sec:measuring}

The evidence across three experiments supports a hybrid view of current LLM benchmark performance. Scores reflect a mixture of: (a)~\emph{large-scale memorization} of question text, answer options, and surrounding online discussion; (b)~\emph{statistical generalization within the training distribution}, enabling correct interpolation between seen examples; and (c)~\emph{variable surface-pattern reliance}, with sensitivity to surface-form changes that correlates with contamination level.

The exam metaphor is instructive. Human exams are designed under the assumption that test items are completely unseen. When an LLM is evaluated on a benchmark it has ingested, the situation is closer to a student sitting an exam after receiving---in some cases---the complete question set itself. A decontaminated, withheld benchmark with surface-form variation testing is the appropriate analogue of a genuine novel exam.

Common evaluation pitfalls include: reporting raw accuracy on publicly available benchmarks without contamination analysis; generalizing from narrow benchmark domains to broad occupational competence; treating single-number accuracy scores as representative of robustness across prompt formulations; and equating average accuracy with fitness for high-stakes deployment where calibration and failure mode matter as much as mean performance.

\section{Normative and Policy Implications}
\label{sec:normative}

\subsection{Disclosure obligations}

If benchmark scores are materially influenced by prior exposure to test items, model providers arguably have an obligation to disclose this. The three-experiment pipeline in this paper demonstrates that contamination analyses are feasible with public infrastructure and modest API costs. Transparent reporting of contamination analyses alongside performance claims would allow downstream users---employers, policymakers, judges, and researchers---to calibrate expectations appropriately.

\subsection{Implications for legal and medical practice}

In legal practice, benchmark scores are sometimes invoked in support of claims about model reliability. Courts and regulators should treat such claims with caution: a model performing well on contaminated law questions may still hallucinate case law, misinterpret statutes, or fail to recognize jurisdictional conflicts~\cite{stanfordlaw}. The 19.8~percentage-point accuracy drop under indirect referencing for Law questions (Experiment~2) indicates that legal reasoning performance on MMLU substantially reflects familiarity with specific question phrasings rather than transferable legal analysis. The same concern applies in medicine, where overreliance on benchmark scores may lead to deployment of systems with brittle, memorization-driven performance in high-stakes clinical contexts.

\subsection{Regulatory evaluation standards}

Contamination-aware evaluation is an essential component of rigorous AI regulation. Regulatory guidance for high-stakes AI systems should require: evaluation on decontaminated or proprietary benchmarks not included in training data; behavioral contamination analyses (TS-Guessing or equivalent); robustness testing across prompt formulations including paraphrase and indirect-reference variants; and standardized disclosure of evaluation methodology. The three-experiment framework presented here provides a replicable, low-cost template for such audits.

\section{Future Directions}
\label{sec:future}

Several extensions follow directly from this work. A properly powered private benchmark comparison---with at least 100--200 carefully difficulty-calibrated questions per domain, authored after model training cutoffs and verified offline---would complement the behavioral evidence here with a direct accuracy measurement on genuinely novel material. Domain-matched difficulty calibration is essential and could be established by piloting questions on a human cohort before LLM evaluation.

The TS-Guessing protocol should be extended to sentence-level reconstructions and multi-hop compositional probes, providing a richer profile of what models store and in what form. The DeepSeek-R1 anomaly---distributed memorization without verbatim recall---deserves dedicated investigation: understanding whether this pattern is architectural (chain-of-thought pretraining depth), corpus-related, or a deliberate training design choice has significant implications for contamination mitigation strategy.

The convergence between Experiments~1 and~3 on category ordering invites a systematic study of which web-presence features best predict behavioral memorization, potentially enabling lightweight contamination screening without requiring full API evaluation of all models.

Finally, the interaction between contamination and hallucination deserves direct causal study using controlled synthetic benchmarks where memorization exposure can be precisely manipulated, allowing clean estimation of how much contamination contributes to both inflated scores and downstream hallucination frequency.

\section{Conclusion}
\label{sec:conclusion}

Returning to the motivating question---are LLMs truly smarter than humans, or did they simply see the exam?---the evidence from three independent experiments points to the same answer: the question cannot be definitively resolved with current benchmark practice, and the available evidence gives substantial reason for skepticism.

Experiment~1 establishes that 13.8\% of MMLU questions are contaminated under a conservative two-condition detection rule, rising to 18.1\% in STEM and 66.7\% in the most affected individual subject. Experiment~2 shows that surface-form changes reduce model accuracy by an average of 7.0 percentage points across all six models, with the sharpest degradation in the highest-contamination domains (Law: $-$19.8~pp, Ethics: $-$19.8~pp)---direct behavioral evidence that a meaningful portion of MMLU performance reflects surface-pattern familiarity rather than transferable knowledge. Experiment~3 finds that 72.5\% of MMLU questions trigger memorization signals far above random chance across six frontier models, with DeepSeek-R1 displaying a distinctive distributed memorization signature that explains its anomalous Experiment~2 profile.

All three experiments agree on the category-level ordering of contamination severity: STEM $>$ Professional $>$ Social Sciences $>$ Humanities. This convergence across three methodologically independent approaches---external web detection, accuracy degradation under surface-form perturbation, and internal reconstruction probing---constitutes the strongest available multi-method evidence that MMLU contamination is a real, structural, and practically consequential phenomenon.

The three-experiment framework presented here is fully replicable with public infrastructure at modest cost and should be adopted as a standard audit protocol. Claims of human-level or superhuman AI performance derived from contaminated public benchmarks deserve the same scrutiny we would apply to any scientific claim where the evaluation procedure is known to be compromised.

\end{document}